\documentclass[11pt]{article}

\usepackage{fontspec}
\usepackage{xeCJK}
\usepackage{amsmath}
\usepackage{amssymb}
\usepackage{svg}
\usepackage{xcolor}
\usepackage{tcolorbox}
\usepackage{float}
\usepackage{geometry}
\usepackage{setspace}

\usepackage{natbib}

\usepackage{acl}

\usepackage[english,bidi=default]{babel} 
\babelfont{rm}{TeXGyreTermesX} 

\usepackage{latexsym}
\usepackage{bm}
\usepackage{xcolor}

\usepackage{microtype}

\usepackage{graphicx}
\usepackage{booktabs}
\usepackage{multirow}
\usepackage{tabularx}
\usepackage{booktabs}
\usepackage{algorithmic}
\usepackage{algorithm}

\usepackage{hyperref}
\usepackage{colortbl}

\title{ToxiTrace: Gradient-Aligned Training for Explainable \\ Chinese Toxicity Detection}

\author{
 \textbf{Boyang Li}$^{1}$, 
 \textbf{Hongzhe Shou}$^{1}$, 
 \textbf{Yuanyuan Liang}$^{2}$, 
 \textbf{Jingbin Zhang}$^{1}$, 
 \textbf{Fang Zhou}$^{1}$\thanks{Corresponding author.} \\
    $^{1}$School of Data Science and Engineering, East China Normal University \\
    $^{2}$School of International Chinese Studies, East China Normal University \\
    {\ttfamily \{byli1024, hzshou, jingbinzhang\}@stu.ecnu.edu.cn} \\
    {\ttfamily yyliang@chinese.ecnu.edu.cn, fzhou@dase.ecnu.edu.cn}
}

\begin{document}

\maketitle

\begin{abstract}

Existing Chinese toxic content detection methods mainly target sentence-level classification but often fail to provide readable and contiguous toxic evidence spans. We propose \textbf{ToxiTrace}, an explainability-oriented method for BERT-style encoders with three components: (1) \textbf{CuSA}, which refines encoder-derived saliency cues into fine-grained toxic spans with lightweight LLM guidance; (2) \textbf{GCLoss}, a gradient-constrained objective that concentrates token-level saliency on toxic evidence while suppressing irrelevant activations; and (3) \textbf{ARCL}, which constructs sample-specific contrastive reasoning pairs to sharpen the semantic boundary between toxic and non-toxic content. Experiments show that ToxiTrace improves classification accuracy and toxic span extraction while preserving efficient encoder-based inference and producing more coherent, human-readable explanations. We have released the model at \href{https://huggingface.co/ArdLi/ToxiTrace}{https://huggingface.co/ArdLi/ToxiTrace}.

\end{abstract}

\vspace{6pt}
\noindent
\textcolor{red}{\textbf{Disclaimer:}\quad \textit{This paper describes violent and discriminatory content that may be disturbing to some readers.}}

\section{Introduction}
In the era of pervasive digital social media, toxic user-generated content (UGC)—such as cyberbullying and hate speech—has become increasingly prevalent, posing tangible risks to online communities and society at large. As a result, toxic content detection has been extensively studied~\cite{arora2023zongshu,kirk-etal-2022-zongshu,azumah2023zongshu}, with Transformer-based pre-trained language models (PLMs)~\cite{NIPS2017_attention,devlin-etal-2019-bert,liu2019roberta} and, more recently, large language models (LLMs) further advancing classification performance.

\begin{figure}[!t]
    \centering
    \includegraphics[width=1.0\linewidth]{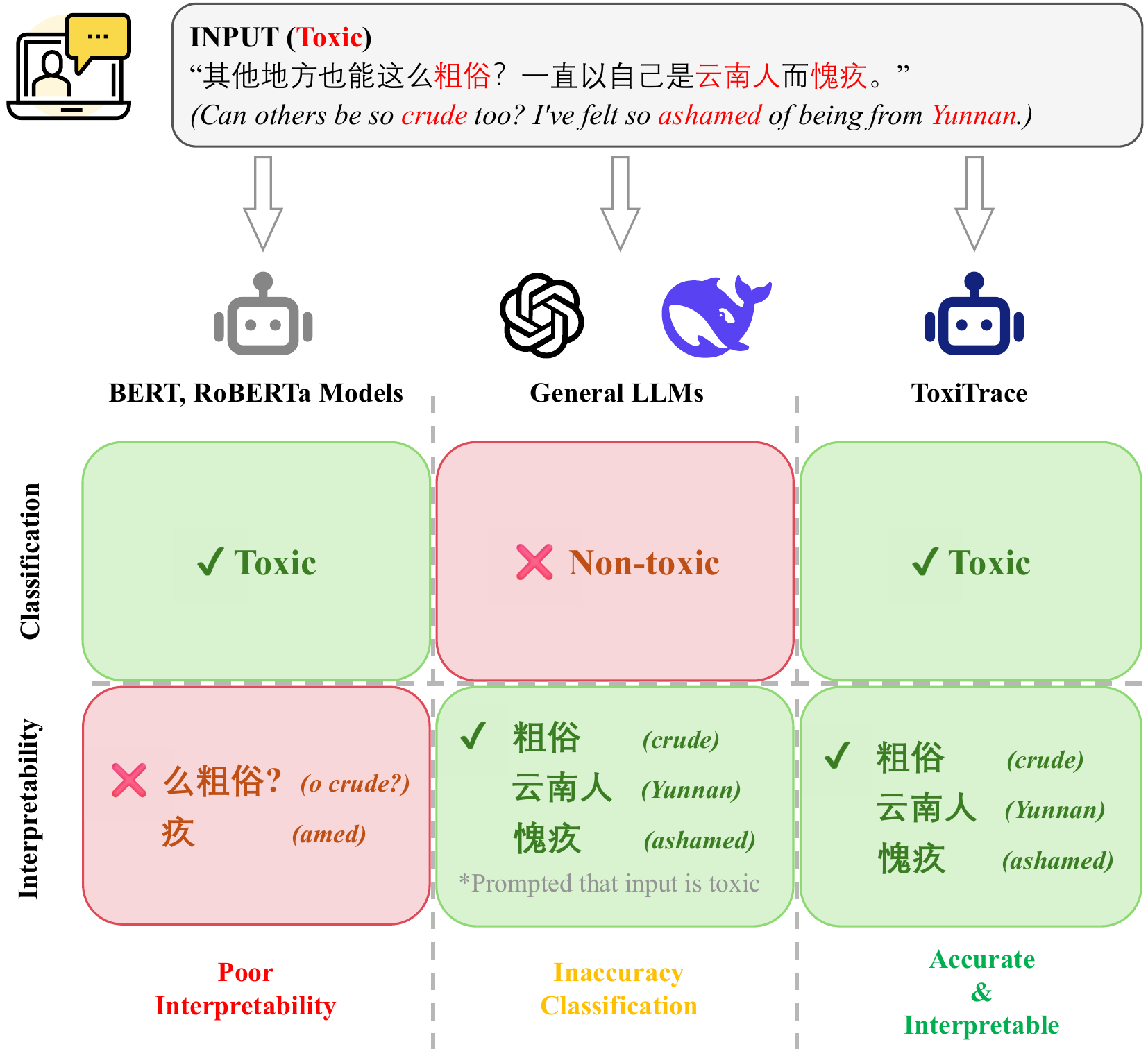}
    \caption{Existing encoder-based detectors struggle to reliably extract fine-grained toxic expressions within a sentence; LLMs can better extract spans when toxicity is given but are limited in direct classification and efficiency; our method preserves classification performance while enabling contiguous toxic span extraction.
    }
    \label{fig:sample}
    \vspace{-1em}
\end{figure}

Despite these advances, most existing work focuses on sentence-level toxicity classification, while providing little insight into which specific parts of a sentence constitute the toxic content. However, identifying fine-grained toxic evidence is crucial for explainability, moderation transparency, and downstream interventions. This challenge is particularly pronounced for Chinese.

Unlike English, where words serve as the basic semantic units, Chinese toxic expressions are typically realized as multi-character phrases, while individual characters are often semantically ambiguous. However, mainstream Chinese PLMs adopt character-level tokenization~\cite{cui-etal-2020-chinese-roberta,sun-etal-2021-chinesebert}. Consequently, attribution signals such as gradients or attention weights are fragmented across individual characters rather than coherent semantic spans, producing rationales that are difficult for humans to interpret~\cite{tiDuFangFa_nlp}.

As a result, existing Chinese toxic content detection approaches—despite achieving strong sentence-level performance through fine-tuning~\cite{deng-etal-2022-cold}, knowledge distillation~\cite{augCold}, or glyph-aware modeling~\cite{Character-level_hyper,jieHeZiXing_1}—remain limited in their ability to accurately extract the true toxic expressions within a sentence (Figure~\ref{fig:sample}, left). In contrast, LLMs often exhibit stronger capabilities in explanation and span extraction~\cite{LLM_JieShiXing,LLM_JieShiXing_2}, but they typically underperform on direct toxicity classification and incur substantially higher inference costs~\cite{k2024-llm_tox_detection,LLM_classification_normal,YouHaiWenBen_LLMbad} (Figure~\ref{fig:sample}, middle).

\begin{figure*}[ht]
    \centering
    \includegraphics[width=1.0\linewidth]{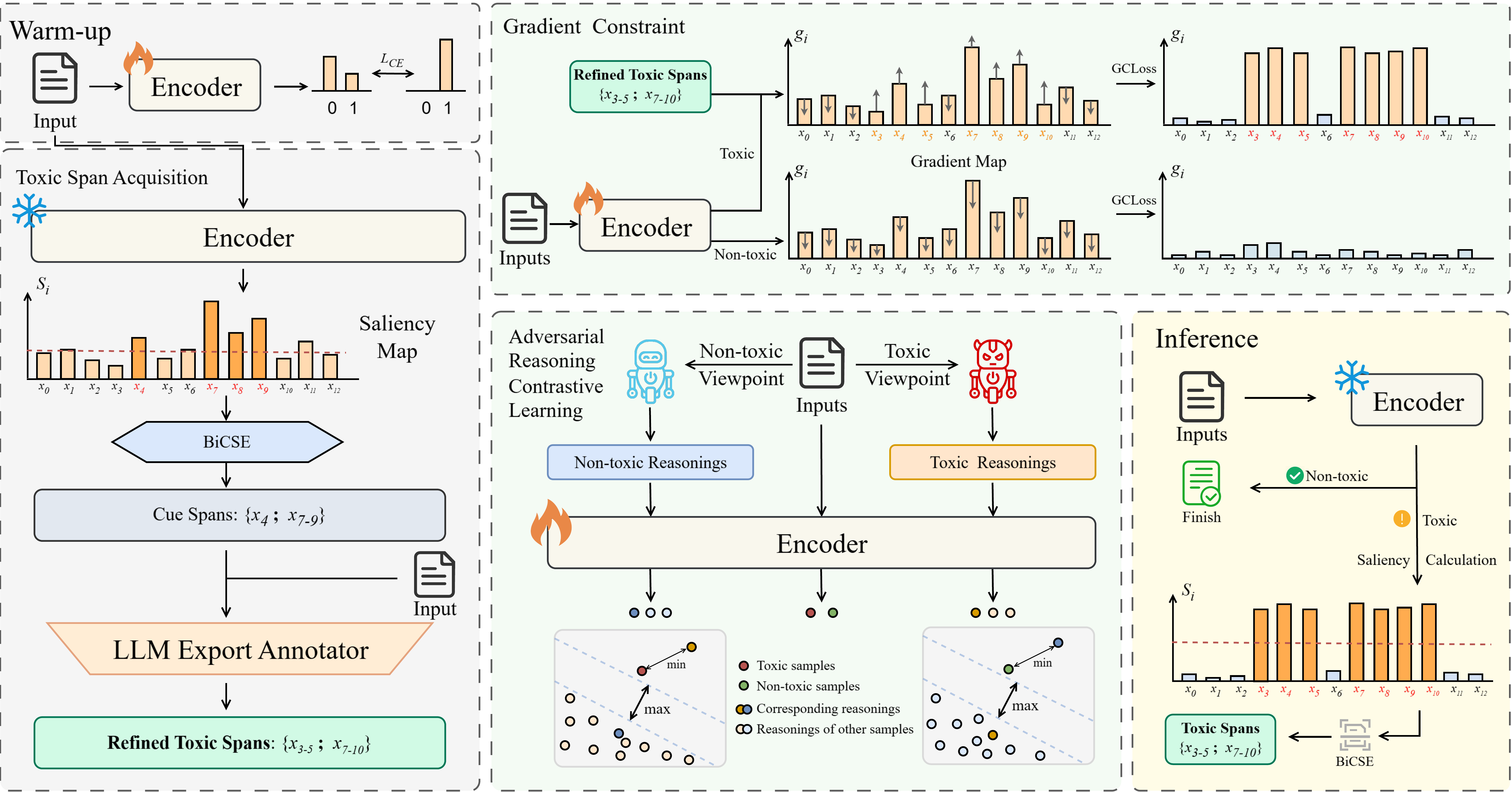} \\
    \caption{Framework of the proposed ToxiTrace method. During training, we warm up an encoder classifier, acquire weak span annotations with CuSA using BiCSE-extracted saliency cues, and jointly optimize \textsc{GCLoss} and ARCL to concentrate saliency on toxic evidence. During inference, the model predicts toxicity and, for toxic inputs, extracts contiguous spans via BiCSE from the saliency map.}
    \label{架构图}
    \vspace{-1em}
\end{figure*}

These limitations call for a method that preserves the classification strength and efficiency of encoder-based models while enabling reliable extraction of contiguous, human-readable toxic spans (Figure~\ref{fig:sample}, right). To this end, we propose \textbf{ToxiTrace}, a span-extraction-oriented framework built on BERT-style encoders for Chinese toxic content detection, designed to produce coherent and interpretable toxic rationales without requiring fine-grained span supervision. Our main contributions are as follows:

\begin{itemize}
    \item [$\bullet$]  We propose a cue-guided span annotation strategy with gradient-aware training, which leverages attribution signals from encoders to induce consistent saliency on toxic tokens without requiring explicit span annotations.
    
    \item [$\bullet$]  We introduce a bidirectional cliff-based span extraction algorithm to identify contiguous toxic spans based on saliency transitions, alleviating the span fragmentation issue inherent in prior top-$k$ selection methods.
    
    \item [$\bullet$]  We develop an adversarial reasoning contrastive learning objective with adaptive InfoNCE, which aligns sample-specific toxic and non-toxic reasoning representations, sharpening the semantic boundary and further enhancing span-level interpretability.

     \item [$\bullet$] Experiments on multiple Chinese toxic content benchmarks demonstrate that \textbf{ToxiTrace} consistently outperforms strong baselines.
\end{itemize}

\section{Related Work}

\subsection{Toxic Content Detection}
Toxic content detection has evolved from early Bag-of-Words and conventional classifiers~\cite{Kwok_Wang_2013_cidai,waseem-hovy-2016-chuantongML,Davidson_2017_chuantongML} to neural and Transformer-based models that achieve strong sentence-level accuracy~\cite{deeplearning,zimmerman-etal-2018-DeepLearning,caselli-etal-2021-hatebert,sarkar-etal-2021-fbert-neural}. 

Chinese toxic content detection follows a similar trajectory, supported by datasets such as COLD~\cite{deng-etal-2022-cold}, ToxiCN~\cite{lu-etal-2023-facilitating}, and CNTP~\cite{yang-etal-2025-CNTP}, as well as encoder-centric methods including character-level modeling~\cite{Character-level_hyper}, domain feature fusion~\cite{domain_chinese}, LLM-assisted rewriting~\cite{LLMjian_ce}, and distillation for robustness~\cite{augCold}. 
In contrast to prior work that primarily optimizes sentence-level detection, we target fine-grained toxic span extraction by training an encoder to produce evidence-aligned saliency and readable spans.

Although CNTP~\cite{yang-etal-2025-CNTP} provides limited span annotations, most Chinese detection pipelines still lack reliable supervision and methods for extracting \emph{contiguous} toxic spans within sentences. This gap often leaves models accurate yet poorly grounded in human-readable evidence. 
We address this by using cue-guided span signals to support training under weak supervision and by enforcing higher gradient responses on toxic evidence.

\subsection{Attribution Method}
Attribution methods, initially developed in computer vision, have been widely adopted for NLP interpretability. Representative lines include perturbation-based explanations such as LIME~\cite{2016_lime}, gradient-based explanations~\cite{Right_for_2017}, and task-calibrated attention-saliency alignment that improves faithfulness~\cite{enjoy_sal,Attention-based_Explanations}. A practical issue is that many pipelines extract explanations by selecting top-$k$ salient tokens, where $k$ is either fixed~\cite{fixed_k_1,fixed_k_2} or set by heuristics such as a fixed fraction of sentence length~\cite{fixed_fraction}; dynamic alternatives such as peak-based top-$k$ have been proposed to improve consistency~\cite{kamp-etal-2023-topk}. 
Building on gradient-based attribution, we explicitly \emph{train} the encoder to yield more evidence-aligned gradients, rather than only post-hoc selecting salient tokens.

Despite progress, top-$k$ selection often produces fragmented highlights and cannot recover contiguous, human-readable spans—an issue that is especially pronounced under character-level tokenization. 
We therefore propose a bidirectional scanning algorithm that identifies consecutive locally high-saliency spans, enabling stable contiguous toxic span extraction beyond discrete token selection.

\section{Methodology}

As shown in Figure~\ref{架构图}, our ToxiTrace framework contains following four steps: 
(1) We first warm up the encoder with standard classification training to obtain robust sentence-level discrimination. 
(2) After warming-up, we compute saliency maps and apply a \textbf{Bi}directional \textbf{C}liff-based \textbf{S}pan \textbf{E}xtraction algorithm (BiCSE) to obtain initial high-saliency spans, which are used as cues to prompt an LLM to refine boundaries and recover coherent toxic spans, yielding \emph{refined toxic spans} as weak span annotations. 
(3) We then introduce \textbf{G}radient \textbf{C}onstraint \textbf{Loss} (\textsc{GCLoss}) to explicitly increase gradient responses on toxic evidence while suppressing spurious activations on non-toxic tokens, shaping a more concentrated and extractable saliency. 
(4) In parallel, we adopt \textbf{A}dversarial \textbf{R}easoning \textbf{C}ontrastive \textbf{L}earning (ARCL) with adaptive InfoNCE to align each input with sample-specific reasoning of opposing stances, sharpening the toxic/non-toxic semantic boundary. 

At inference time, the model first predicts toxicity; if toxic, the saliency map will be calculated, and toxic spans will be extracted using BiCSE.

\subsection{Cue-guided Span Annotation (CuSA)}
Existing toxic content datasets only have coarse-grained labels (toxic/non-toxic) and cannot accurately locate toxic spans. 
CuSA constructs span-level signals by using the model's attribution map as cues and letting an LLM refine span boundaries.
It consists of two steps: (1) warm-up fine-tuning to obtain a reliable sentence-level classifier; and (2) cue-guided span refinement, where we feed the toxic text together with initially extracted salient spans as cues to an LLM for span annotation.

\paragraph{Warm-up training.}
We train the encoder with binary cross-entropy. Given an input text sequence $\mathcal{X} = \{x_1, \ldots, x_n\}$, the embedding layer maps tokens to $\mathcal{E}=[\boldsymbol{e}_1, \ldots, \boldsymbol{e}_n]$, the model $\phi$ generate its contextual representation $\mathcal{H}=\phi(\mathcal{E})=[\boldsymbol{h}^{cls}, \boldsymbol{h}^{1}, \boldsymbol{h}^{2}, ... \boldsymbol{h}^{n}]$. This representation is then fed into a classification head $\psi$ to yield the predicted probability $P(y|\mathcal{X})=\psi(\boldsymbol{h}^{cls})$.

\paragraph{Saliency cues for span annotation.}
After warm-up, following existing attribution methods~\cite{enjoy_sal,gradient_integrated}, we compute a token-level \emph{saliency} score for each token and form a saliency map, which serves as cues for span extraction.
The saliency $s_i$ of each token $x_i$ can be calculated via Eq.~\eqref{sal_i}:
\begin{equation}
\label{sal_i}
s_i = \left\lVert \boldsymbol{e}_i \odot \frac{\partial \log P(y|\mathcal{X})}{\partial \boldsymbol{e}_i} \right\rVert_2.
\end{equation}

Existing attribution methods select the top-$k$ most salient tokens~\cite{kamp-etal-2023-topk}; however, these selected tokens tend to be scattered. To address this limitation, we propose BiCSE 
(detailed in Appendix~\ref{algorithm}),
a bidirectional cliff-based scanning algorithm that tracks saliency transitions to identify the optimal start and end boundaries of contiguous spans. Moreover, longer and more continuous toxic spans are captured by taking the union of results from two sequential scans (left-to-right and right-to-left).

To help the LLM to find more potential toxic spans, both the raw text and the initially extracted spans (cues) $\mathcal{T}_{cue}$ are fed into the LLM. In our experiments, we use Gemini 2.5 Pro~\cite{geminiteam2025geminifamilyhighlycapable} as an expert annotator to integrate and refine these cues. The refined toxic spans are formulated as: $\mathcal{T}_{refined} = \text{LLM}(\boldsymbol{x}, \mathcal{T}_{cue})$. By leveraging the LLM's superior interpretability, we can achieve more accurate annotation of toxic spans. 

\subsection{Gradient Constraint Loss}
The fine-tuned model attains satisfactory classification performance overall; yet,
its token-level toxicity discrimination remains imprecise (Appendix~\ref{change_of_sal} shows the saliency maps before and after applying our ToxiTrace method). 
CuSA provides refined toxic spans as annotations, which allow us to explicitly shape the model's token-level attribution for span extraction.
Concretely, \textsc{GCLoss} consists of two complementary components: (i) a \textbf{Pairwise Gradient Ranking (PGR)} term that enforces a margin between toxic and non-toxic tokens, and (ii) a \textbf{Push--Pull Threshold (PPT)} term that regularizes their gradient ranges within each sample.
During training, both terms operate on the gradient norm of the log-predicted probability with respect to the \emph{input embeddings} $\boldsymbol{e}_i$:
\begin{equation}
\label{g_i1}
g_i=\left\lVert \frac{\partial \log P(y|\mathcal{X})}{\partial \boldsymbol{e}_i} \right\rVert_2
\end{equation}
as the training signal to increase responses on toxic spans and suppress spurious activations on non-toxic tokens, thereby shaping more concentrated and extractable attribution. 

\paragraph{PGR Loss.} 
The objective of this loss is to penalize cases where, within a single sentence, the gradient of a toxic token exceeds that of a non-toxic token by a margin smaller than the predefined threshold $m$. The specific formula is given as follows:
\begin{equation}\mathcal{L}_{PGR} = \frac{1}{|\mathcal{P}| \cdot |\mathcal{N}|} \sum_{i \in \mathcal{P}} \sum_{j \in \mathcal{N}} \max(0, g_j - g_i + m),
\end{equation}
\noindent where $\mathcal{P}$ and $\mathcal{N}$ denote the sets of toxic and non-toxic tokens, respectively; $g_{i}$, $g_{j}$ is calculated according to Eq. \eqref{g_i1}; and $m$ is set to 1 in this study.

\paragraph{PPT Loss.}
The PGR Loss introduced above captures the relative gradient relationship between toxic and non-toxic tokens, yet it cannot constrain the gradient value ranges of either token type. Given the substantial discrepancies in gradient scores across different sentences, employing a fixed range to regulate the gradients of toxic and non-toxic tokens is inherently flawed. To address this dual limitation, we propose the intra-sentence PPT Loss that leverages gradient information of tokens within each single sample to separately guide the gradient learning of toxic and non-toxic tokens. Specifically, we first calculate the 15th percentile of the gradient values of all tokens in a single sample as the threshold $\tau$, which serves to constrain the gradient values of non-toxic tokens to stay below this threshold. The detailed formula is given as follows:

\begin{equation}
    \mathcal{L}_{neg} = \frac{1}{|\mathcal{N}|} \sum_{j \in \mathcal{N}} [\max(0, g_j - \tau)].
\end{equation}

For toxic tokens, we expect their gradient values to fall within a relatively high range. 
Thus, we adopt the maximum gradient $g_{\text{max}}$ as the reference and set $\alpha \cdot g_{\text{max}}$ as the lower bound, where $\alpha$ denotes a positive target coefficient. 
Our goal is to push the gradient values of toxic tokens above this threshold. However, given the possibility of an excessively large $g_{\text{max}}$, we further introduce a gradient cap $\tau_{\text{cap}}$ to prevent gradient explosion caused by overly high gradient values of toxic tokens. The formula for toxic tokens is given as follows:

\begin{equation}
    \mathcal{L}_{pos} = \frac{1}{|\mathcal{P}|} \sum_{i \in \mathcal{P}} [\max(0, t_n - g_i)],
\end{equation}

\noindent where the target value $t_n = \min(\alpha \cdot g_{max}, \tau_{cap})$. 
The overall form of the PPT Loss is:

\begin{equation}
    \mathcal{L}_{PPT} = \frac{1}{2} (\mathcal{L}_{pos} + \mathcal{L}_{neg})
\end{equation}

\noindent\textbf{Remark.}
\textsc{GCLoss} constrains \emph{gradients} during training (Eq.~\eqref{g_i1}), because gradients directly capture the model's sensitivity to token-level evidence.
For span extraction at inference time, we compute a \emph{saliency} score with the same embedding-level gradients.

This formulation reflects both sensitivity and contribution, and BiCSE is applied to the saliency sequence $\{s_i\}_{i=1}^n$ to produce contiguous spans.

\begin{table*}[t]
\centering
\scalebox{0.64}{
\begin{tabular}{lcccccccccc}
\toprule
\multirow{2}{*}{\bf Models} &
\multicolumn{5}{c}{\bf COLD} &
\multicolumn{5}{c}{\bf ToxiCN} \\
\cmidrule(lr){2-6} \cmidrule(lr){7-11}
& \bf $Acc$ & \bf $R$ & \bf $P$ & \bf $F_1$ & \bf Macro-$F_1$
& \bf $Acc$ & \bf $R$ & \bf $P$ & \bf $F_1$ & \bf Macro-$F_1$ \\
\midrule

Qwen3-8B & $74.03_{0.23}$ & $59.90_{0.28}$ & $70.15_{0.32}$ & $64.62_{0.19}$ & $72.33_{0.21}$ 
& $71.44_{0.33}$ & $82.94_{0.29}$ & $69.13_{0.27}$ & $75.41_{0.16}$ & $71.51_{0.12}$ \\
LLaMA3.1-8B & $73.91_{0.31}$ & $51.46_{0.29}$ & $74.58_{0.35}$ & $60.90_{0.22}$ & $72.00_{0.28}$ 
& $65.18_{0.28}$ & $44.29_{0.29}$ & $81.61_{0.24}$ & $57.42_{0.20}$ & $68.25_{0.22}$ \\
DeepSeek-V3.2 & $74.23_{0.16}$ & $59.23_{0.10}$ & $70.91_{0.13}$ & $64.55_{0.12}$ & $72.51_{0.09}$ 
& $60.66_{0.11}$ & $36.19_{0.14}$ & $77.35_{0.16}$ & $49.30_{0.08}$ & $64.14_{0.11}$ \\
GPT-4o & $74.49_{0.18}$ & $67.79_{0.20}$ & $66.98_{0.21}$ & $67.38_{0.12}$ & $73.22_{0.13}$ 
& $73.20_{0.21}$ & $65.66_{0.18}$ & $66.98_{0.16}$ & $67.38_{0.11}$ & $73.22_{0.09}$ \\
\midrule 
RoBERTa & $82.68_{0.42}$ & $86.37_{0.86}$ & $74.42_{0.86}$ & $79.80_{0.27}$ & $82.56_{0.38}$
& $82.70_{0.45}$ & $84.38_{0.30}$ & $83.40_{0.55}$ & $83.89_{0.40}$ & $82.81_{0.45}$ \\
Qwen3-8B (SFT) & $82.25_{0.22}$ & $82.68_{0.22}$ & $75.02_{0.38}$ & $78.66_{0.16}$ & $81.87_{0.10}$ 
& $82.08_{0.54}$ & $83.52_{1.33}$ & $82.74_{0.93}$ & $83.12_{0.52}$ & $82.02_{0.42}$ \\
LLaMA2-7B (SFT) & $82.43_{0.28}$ & $86.28_{0.18}$ & $73.78_{0.22}$ & $79.54_{0.10}$ & $82.46_{0.13}$ 
& $81.25_{0.37}$ & $84.62_{0.44}$ & $80.81_{0.40}$ & $82.67_{0.29}$ & $81.18_{0.22}$ \\
LLaMA3.1-8B (SFT) & $83.00_{0.10}$ & \underline{$87.13_{0.16}$} & $74.37_{0.18}$ & $80.19_{0.09}$ & $82.19_{0.10}$ 
& $83.02_{0.17}$ & $86.08_{0.20}$ & $82.52_{0.24}$ & $84.26_{0.12}$ & $82.96_{0.14}$ \\
\midrule 
LLaMA3.1-8B + ToxiTrace & $82.15_{0.24}$ & $\boldsymbol{88.42}_{0.32}$ & $72.52_{0.37}$ & $79.68_{0.20}$ & $81.89_{0.27}$ 
 & $82.33_{0.47}$ & $84.93_{0.53}$ & $82.22_{0.42}$ & $83.55_{0.35}$ & $82.23_{0.32}$ \\
Qwen3-8B + ToxiTrace & $82.96_{0.29}$ & $82.49_{0.37}$ & $\boldsymbol{76.38}_{0.41}$ & $79.41_{0.25}$ & $82.39_{0.21}$ 
 & $82.91_{0.62}$ & $83.44_{0.85}$ & $\boldsymbol{84.10}_{0.77}$ & $83.77_{0.31}$ & $82.86_{0.36}$ \\
RoBERTa + ToxiTrace & \bf $\boldsymbol{83.84}_{0.27}$ & ${86.19}_{0.48}$ & \underline{${76.14}_{0.34}$} & \bf $\boldsymbol{80.85}_{0.14}$ & \bf $\boldsymbol{83.68}_{0.16}$ 
& \underline{$83.62_{0.15}$} & \underline{$87.05_{1.28}$} & \underline{$82.82_{0.78}$} & \underline{$84.88_{0.27}$} & \underline{$83.56_{0.14}$} \\
MacBERT + ToxiTrace & \underline{$83.22_{0.19}$} & $86.19_{0.48}$ & $75.10_{0.49}$ &\underline{$80.27_{0.10}$} & \underline{$83.13_{0.17}$} 
& $\boldsymbol{83.87}_{0.12}$ & $\boldsymbol{87.99}_{0.83}$ & $82.61_{0.41}$ & $\boldsymbol{85.21}_{0.20}$ & $\boldsymbol{83.83}_{0.14}$ \\
\bottomrule
\end{tabular}}
\caption{The average classification results of different models on COLD and ToxiCN datasets(\%) across 3 independent runs. Subscripts denote standard deviations.} 
\label{all performance}
\vspace{-0.5em}
\end{table*}

\subsection{Adversarial Reasoning Contrastive Learning}
\label{3.3}

While the aforementioned loss functions address token-level gradient constraints, they are confined to individual sentences and cannot capture the semantic boundary differentiating toxic from non-toxic sentences. Inspired by the work of \cite{pmlr-v258-rusak25a-infoNCE}, we adopt an adaptive InfoNCE loss to implement semantic contrastive learning. Existing data augmentation methods~\cite{zengQiang_1,zengQiang_2,zengQiang_3} primarily rely on substituting, modifying or add noise to certain words in a sentence. Such operations lack a thorough understanding of the sentence's inherent semantics. To address this limitation, we propose leveraging the \textbf{LLM debate mechanism}~\cite{LLM_debate} to generate \textbf{adversarial reasoning} as augmented samples. 
This approach enables the model to better explore the intrinsic semantic information of sentences and learn to distinguish the differences between toxic and non-toxic texts and sharpening the semantic boundary.

Specifically, we first use two opposing reasoning prompts: \textit{"Assuming the text is \{toxic, normal\}, generate explanations to support this judgment"}\footnote{Detailed prompt templates are provided in Appendix~\ref{appendix_prompt}.}, and feed them to the LLM (Gemini 2.5 Flash) to obtain semantically augmented positive and negative samples. This way, the model can produce targeted reasoning content, instead of generating generic descriptions such as \textit{"This sentence is just an exaggerated joke and does not intend to attack any group"}.

Using the two opposing reasoning prompts obtained above, we adopt the following contrastive loss to facilitate sentence-level semantic learning:

\begin{equation}
    \scalebox{0.75}{$\mathcal{L}_{\{tox,nor\}} = -\log \frac{\exp(\boldsymbol{h}_{t} \cdot \boldsymbol{h}_{\{p, n\}} / \tau)}{\exp(\boldsymbol{h}_{t} \cdot \boldsymbol{h}_{\{p, n\}} / \tau) + \sum_{\boldsymbol{h}^{k}_{\{n,p\}} \in \mathcal{B}} \exp(\boldsymbol{h}_{t} \cdot \boldsymbol{h}^{k}_{\{n,p\}} / \tau)}$},
\end{equation}

\noindent where $\boldsymbol{h}_{t}$, $\boldsymbol{h}_{p}$ and $\boldsymbol{h}_{n}$ denote the semantic embeddings of the target text, its positive reasoning explanation and its negative reasoning explanation, respectively; $\boldsymbol{h}_{p}^{k}$ and $\boldsymbol{h}_{n}^{k}$ denote the final-layer \texttt{[CLS]} token embeddings of the positive and negative reasoning explanations corresponding to other toxic and non-toxic sentences within the training batch $\mathcal{B}$. $\tau$ denotes the temperature parameter, which is set to 0.05 in our experiments.

The overall semantic contrastive loss is defined as the average of these two loss components:

\begin{equation}
    \mathcal{L}_{con} = \frac{1}{2}(\mathcal{L}_{tox} + \mathcal{L}_{nor})
\end{equation}

\subsection{Joint Training Objective}

In the joint training phase, we simultaneously optimize three loss components: the gradient-based binary classification loss, the gradient constraint losses, and the contrastive learning loss. The overall training objective is formulated as:

\begin{equation}
\label{total loss}
\mathcal{L} = \mathcal{L}_{CE} + \lambda_{grad}(\mathcal{L}_{PGR} + \mathcal{L}_{PPT}) + \lambda_{sem}\mathcal{L}_{con},
\end{equation}

\noindent where $\lambda_{grad}$ and $\lambda_{sem}$ are hyperparameters that balance the contributions of the gradient constraint losses and the semantic contrastive loss, respectively.

The overall training pipeline is designed as follows:
(1) First, perform a warm-up training using the cross-entropy loss to equip the model with basic toxicity classification capability.
(2) Subsequently, the \textsc{GCLoss} is introduced to enforce the model to generate higher gradient values for toxic tokens and lower gradient values for non-toxic tokens.
(3) Meanwhile, ARCL is adopted to sharpen the semantic boundary between toxic and non-toxic texts. Detailed training hyperparameters are provided in appendix~\ref{train-detail}.

\section{Experiments}

\subsection{Experimental Setup}
\label{4.1}

\paragraph{Dataset.} 
For the binary toxicity classification task, we evaluate our model on two Chinese datasets: COLD~\cite{deng-etal-2022-cold} (32,480 instances in total, with 5,323 test instances) and ToxiCN~\cite{lu-etal-2023-facilitating} (12,011 instances in total, with 2,411 test instances).

For toxic span extraction, we use the span annotations provided in CNTP~\cite{yang-etal-2025-CNTP} as gold labels, which include 2,533 samples annotated with toxic spans, to assess the model's ability to extract fine-grained toxic expressions within toxic sentences.

\paragraph{Models.} To verify the effectiveness of our proposed method in both binary classification and span-level extraction, we applied our training strategy across different BERT-based models: 
Chinese-RoBERTa-wwm-ext \textbf{(RoBERTa)} and 
Chinese-MacBERT-base \textbf{(MacBERT)}~\cite{cui-etal-2020-chinese-roberta}.
Concurrently, a set of LLMs were chosen to conduct comparative analysis:
\textbf{LLaMA2-7B}, 
\textbf{LLaMA3.1-8B}, 
\textbf{Qwen3-8B}, 
\textbf{DeepSeek-V3} and 
\textbf{GPT-4o}.

\paragraph{Implement.} 
Corresponding models fine-tuned solely on binary classification labels were set as baselines. For open-source LLMs, both their direct inference capabilities and their performance after fine-tuning on the respective datasets (denoted as SFT) were evaluated. Fine-tuning of LLMs was conducted using the open-source toolkit LLaMA-Factory\footnote{https://github.com/hiyouga/LLaMA-Factory} to implement LoRA~\cite{hu2022lora}. Closed-source LLMs were evaluated in a zero-shot setting; the prompt templates used were detailed in Appendix~\ref{appendix_prompt}. Models denoted with "(ours)" were trained using the method proposed in this paper. DeepSeek and GPT-4o were accessed via official APIs, while all other experiments were conducted using four NVIDIA A800 80GB GPUs.

\paragraph{Evaluation metrics.} 
To evaluate toxic content detection performance, we
used five widely adopted metrics: Accuracy ($Acc$), Recall ($R$), Precision ($P$), $F_1$ and Macro-$F_1$ Score.

\subsection{Overall Classification Performance}
\label{4.2}

The classification performance of ToxiTrace and competing methods is summarized in Table~\ref{all performance}. Overall, ToxiTrace achieves the best results across all five metrics on both COLD and ToxiCN. In the zero-shot setting, both open- and closed-source LLMs perform poorly on Chinese toxicity classification (and LLaMA2-7B fails to complete the task due to strict safety alignment), whereas fine-tuning brings LLMs to a level comparable with encoder-based models. In terms of efficiency, encoder models finish inference within 20 seconds on both datasets, while LLMs require 2--9 minutes, reflecting differences in model scale, architecture, and the use of LoRA adapters.


\begin{table*}[t]
\centering
\setlength{\tabcolsep}{13pt}
\scalebox{0.7}{
\begin{tabular}{lcccccccc}
\toprule
\multirow{2}{*}{\bf Models} &
\multicolumn{3}{c}{\bf Overlap} &
\multicolumn{4}{c}{\bf Character-level} &
\multirow{2}{*}{\bf Inference Time} \\
\cmidrule(lr){2-4} \cmidrule(lr){5-8}
& $R$ & $P$ & $F_1$ & $R$ & $P$ & $F_1$ & $IoU$ \\
\midrule
CRF~\cite{crf-2021-lexicon}                & 44.48 & \bf 78.01 & 56.66
& 39.24 & \bf 81.85 & 53.05 & 36.10 & 01m 10s \\
LIME~\cite{2016_lime}               & 41.45 & 35.99 & 38.53
& 81.62 & 33.93 & 47.93 & 31.52 & 27m 20s \\
Attention~\cite{attention-etal-2020-luke}          & 39.01 & 36.15 & 37.52
& 25.43 & 47.68 & 33.17 & 19.88 & 00m 42s \\
IG~\cite{gradient_integrated}                 & 58.30 & 60.01 & 59.14
& 36.99 & 76.46 & 49.86 & 33.21 & About 1.2 hours \\
\midrule
Llama3.2-3B        & 41.03 & 71.02             & 52.01 
& 37.47 & 69.42 & 48.67 & 32.16 & 08m 06s \\
Qwen3-0.6B         & 73.68 & 67.10             & 70.23 
& 73.15 & 62.80 & 68.23 & 51.78 & 09m 20s \\
Qwen3-1.7B         & 77.53 & 71.48 & 74.38 
& 76.01 & 66.72 & 71.29 & 55.11 & 09m 28s \\
Llama-3.1-8B       & 81.32 & 69.21             & 74.78 
& 79.94 & 62.97 & 70.87 & 54.33 & 12m 56s \\
Qwen3-8B           &84.83 & 71.97 & 77.87 
& 90.03 & 63.89 & 74.74 & 59.67 & 14m 33s \\
\midrule
RoBERTa            & 42.37& 68.42 & 52.34 
& 45.68 & 50.16 & 43.02 & 33.63 & 01m 58s \\
RoBERTa*           & 71.27 & 59.87& 65.08 
& 58.29 & 61.82 & 55.20 & 42.86 & 01m 58s \\
RoBERTa + ToxiTrace*    & \textbf{86.36}  & 70.95& \textbf{77.90} 
& \bf 83.53 & 70.06 & \bf 77.63 & \bf 61.56 & 01m 58s \\
\midrule
\rowcolor{cyan!10}
Gemini2.5-Pro      & 86.50 & 75.09             & 80.39 
& 85.98 & 74.24 & 79.67 & 66.22 & About 1.5 hours \\
\bottomrule
\end{tabular}}
\caption{Toxic span extraction performance on CNTP. Results without * correspond to extracting the top-$15\%$ most salient tokens, while results with * use our proposed bidirectional algorithm to extract high-saliency spans. Entries shaded in cyan indicate the results of the "refiner" model used in the CuSA.}
\label{extracting}
\vspace{-1em}
\end{table*}

\subsection{Applicability to LLMs}

Since LLMs are also Transformer-based, we conduct an exploratory study on transferring \mbox{ToxiTrace} to decoder-only LLMs. 
Due to resource constraints, LoRA was applied for parameter-efficient adaptation; we then replaced the decoding head with a binary classification head and optimized the same ToxiTrace objectives.
As shown in Table~\ref{all performance}, ToxiTrace achieves performance comparable to instruction fine-tuning on LLM, yet still falls short of encoder-based models trained with ToxiTrace. 
One possible reason is that LoRA updates only a tiny fraction of parameters (typically $\leq 0.5\%$), limiting its ability to reshape embedding-level gradients distributions required by our gradient-oriented training. 
We leave a systematic study of more effective parameter-efficient gradient shaping for future work.

\subsection{Toxic Span Extraction Performance}
We evaluate toxic span extraction on CNTP using two types of metrics: overlap-based matching between the extracted spans and the ground-truth spans, and character-level Recall, Precision, $F_1$-score, and $IoU$, denoted as \textit{Overlap} and \textit{Character-level} in Table~\ref{extracting}, respectively. For overlap-based evaluation, a prediction is counted as correct if its overlap with the gold span exceeds  50\% (e.g., ground-truth toxic span: "河南人", pred: "河南" is correct, while "南" is incorrect). 

Table~\ref{extracting} reports the extraction results under both evaluation schemes. Here, CRF is trained using LLM-generated span annotations; LIME, high-attention token extraction, and IG are computed on a RoBERTa model trained only with binary classification labels. Under both metrics, compared with existing attribution methods, LLMs as well as our model achieve substantially better span extraction performance. Replacing top-$15\%$ token selection with BiCSE also yields a large gain for RoBERTa (RoBERTa $\rightarrow$ RoBERTa*), confirming the advantage of extracting \emph{contiguous} spans. 
Training with ToxiTrace further improves RoBERTa* to RoBERTa+ours*, yielding an additional 12\% improvement in $F_1$-score under both evaluation schemes and a 19\% gain in character-level $IoU$. These results indicate that CuSA and the gradient-shaping objectives (\textsc{GCLoss}/ARCL) produce saliency that is better aligned with the underlying evidence for extraction.

Across LLMs, extraction generally improves with model size, but RoBERTa+ours* achieves comparable or better $F_1$ than the best LLM (Qwen3-8B) while requiring only about $1/7$ of its inference time, demonstrating a substantially better accuracy--efficiency trade-off, the detailed prompt template is provided in Appendix \ref{appendix_prompt}.

\begin{figure}[!t]
    \centering
    \includegraphics[width=0.48\textwidth]{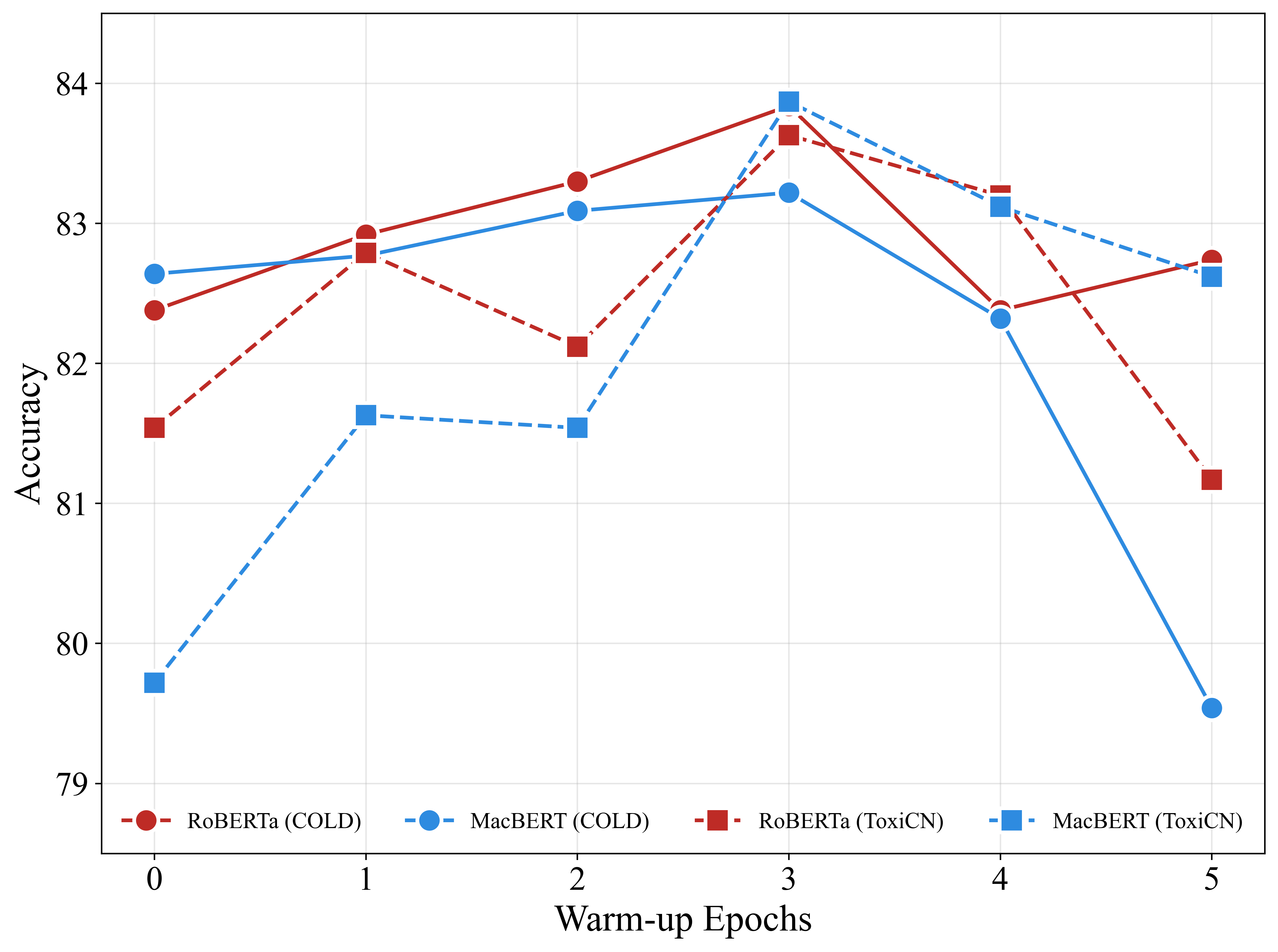} 
    \caption{Final Accuracy with different Warm-up Epochs. The models achieve optimal classification performance when the warm-up steps are set to 3.}
    \label{fig:accuracy_curve}
    \vspace{-0.5em}
\end{figure}

\begin{table*}[t]
\centering
\scalebox{0.8}{
\begin{tabular}{lccccccccc}
\toprule
\multirow{2}{*}{\textbf{Models}} &
\multicolumn{5}{c}{\textbf{Classification}} &
\multicolumn{3}{c}{\textbf{Extraction}} \\
\cmidrule(lr){2-6} \cmidrule(lr){7-9}
& \textbf{$Acc$} & \textbf{$R$} & \textbf{$P$} & \textbf{$F_1$} & \textbf{Macro-$F_1$}
& \textbf{$R$} & \textbf{$P$} & \textbf{$F_1$} \\
\midrule
ToxiTrace & \textbf{83.84} & 86.19 & 76.14 & \bf{80.85} & \bf{83.68} 
& \textbf{86.36} & 70.95 & \bf{77.90} \\
w/o CuSA & 83.26 & 83.77 & \textbf{76.27} & 79.85 & 82.90 
& 65.61 & \bf 79.67 & 71.96 \\
w/o ARCL & 83.13 & 86.71 & 74.72 & 80.27 & 83.12 
& 82.55& 68.99  & 75.16 \\
w/o \textsc{GCLoss} & 83.20 & \textbf{88.04} & 74.29 & 80.58 & 83.36 
 & 75.89& 57.07 & 65.15 \\
RoBERTa & 82.75 & 86.43 & 74.24 & 79.87 & 82.76
& 71.27& 59.87  & 65.08 \\

\bottomrule
\end{tabular}}
\caption{Ablation study results. "Classification" denotes results for the binary classification task on the COLD dataset; "Extraction" denotes results using the toxic span annotations in CNTP~\cite{yang-etal-2025-CNTP} as ground truth labels.}
\label{ablation}
\vspace{-0.5em}
\end{table*}

\begin{figure}[t]
    \centering
    \includegraphics[width=1.0\linewidth]{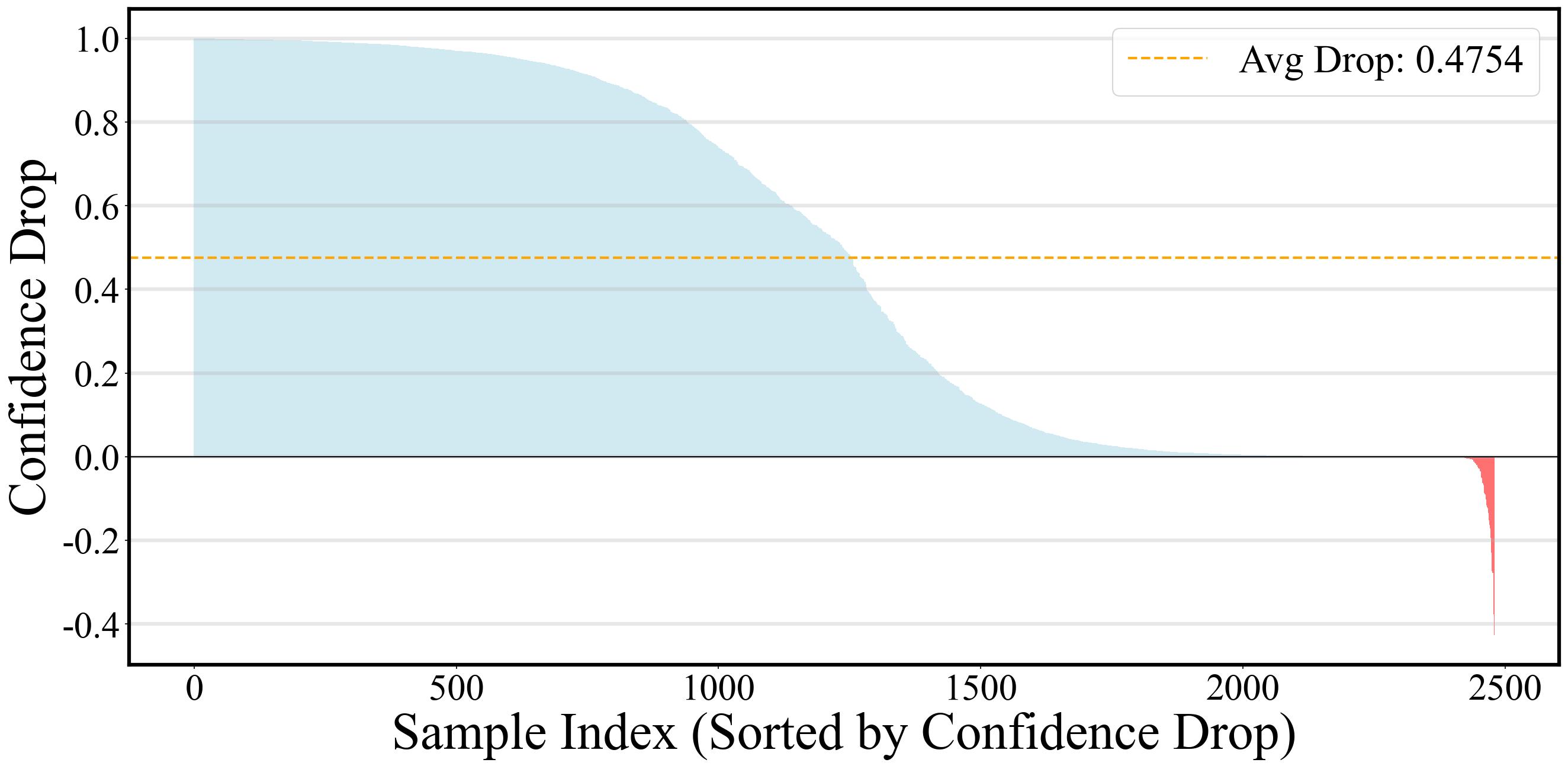}
    \caption{Confidence drop after masking BiCSE-extracted toxic spans for the RoBERTa baseline.}
\label{confidence_drop_1}
\end{figure}

\begin{figure}[t]
    \centering
    \includegraphics[width=1.0\linewidth]{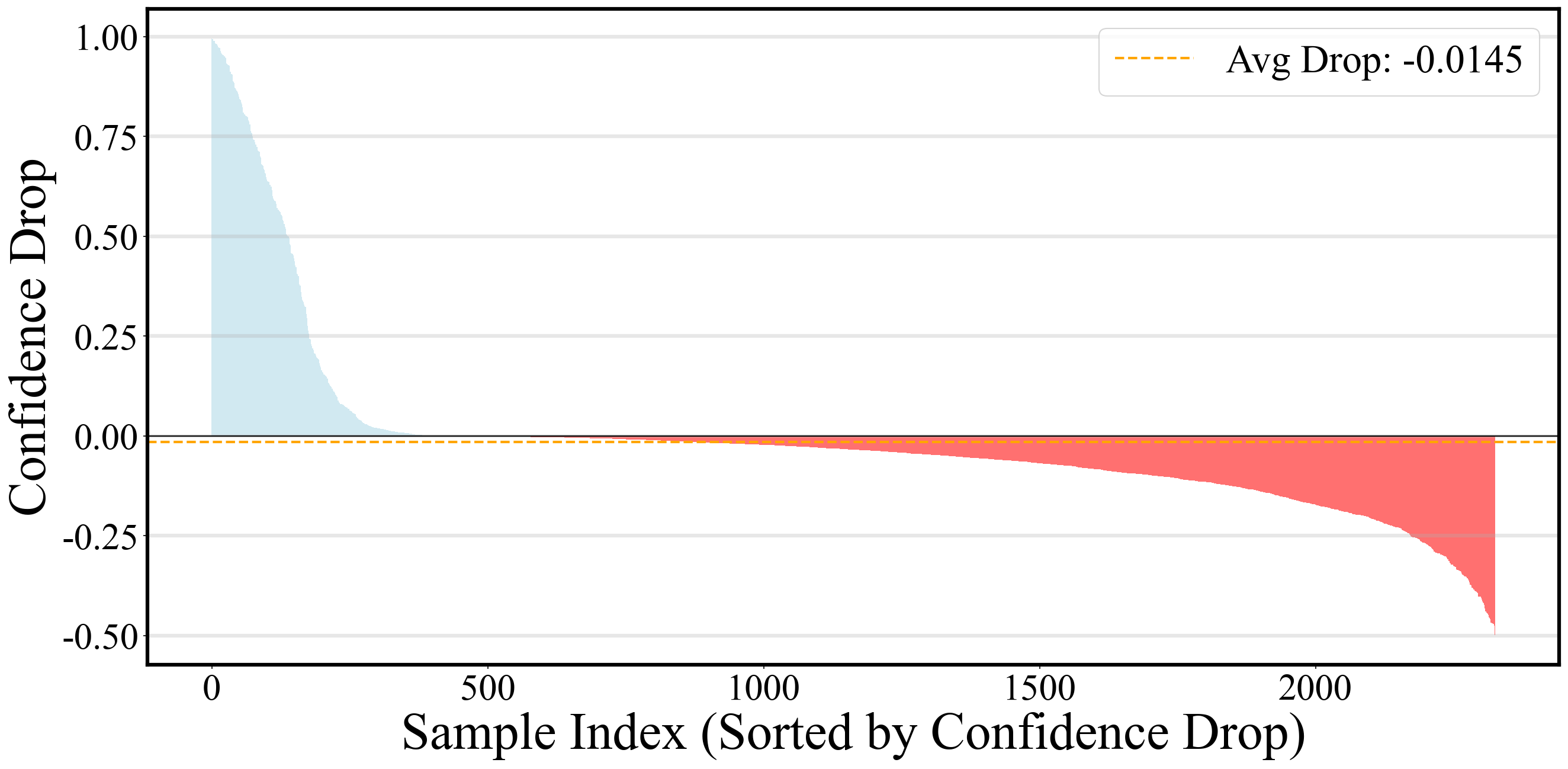}
    \caption{Confidence drop with random masking that matches the number of tokens extracted by BiCSE for RoBERTa trained with \mbox{ToxiTrace}.}
\label{confidence_drop_0}
\end{figure}

\begin{figure}[t]
    \centering
    \includegraphics[width=1.0\linewidth]{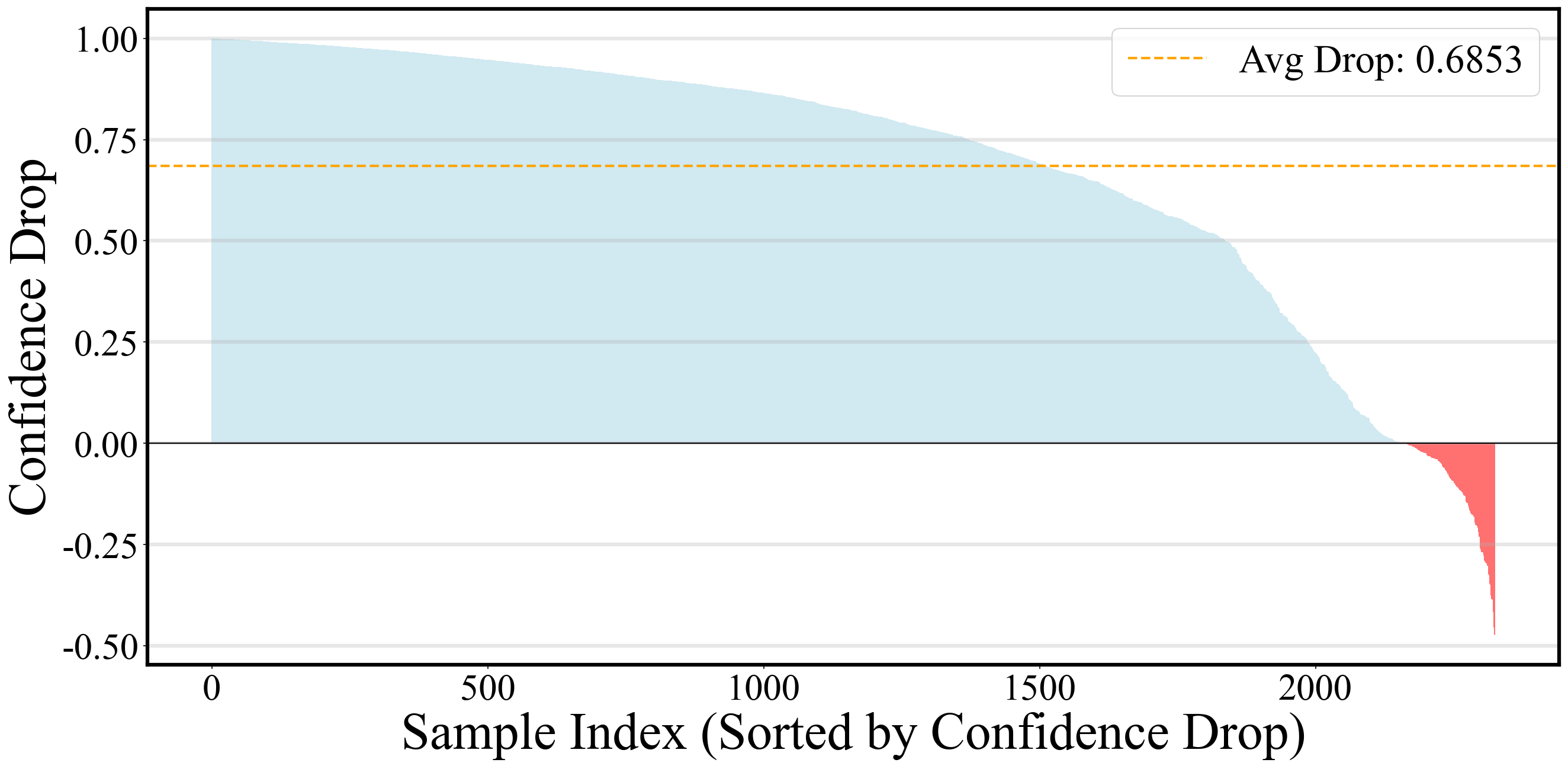}
    \caption{Confidence drop after masking BiCSE-extracted toxic spans for RoBERTa trained with \mbox{ToxiTrace}.}
\label{confidence_drop_ours}
\vspace{-0.5em}
\end{figure}

\subsection{Ablation Study}

We investigate the contributions of the key components in ToxiTrace through an ablation study with three variants, as shown in Table~\ref{ablation}. 
Specifically, \textit{Full} denotes the complete model that jointly optimizes \textsc{GCLoss} and ARCL, while \textit{w/o CuSA} meaning BiCSE cues only, without any LLM refinement, \textit{w/o ARCL} and \textit{w/o \textsc{GCLoss}} remove the adversarial reasoning contrastive objective and the gradient constraint loss, respectively. 
We further include the vanilla \textit{RoBERTa} as the baseline.

Since the ablation trends are consistent across COLD and ToxiCN, we only report results on COLD due to space constraints.
Overall, removing either module leads to performance degradation in both classification and extraction. 
For classification, removing ARCL reduces Macro-$F_1$ by $0.56\%$ (and slightly lowers $F_1$), indicating that ARCL provides a consistent gain via semantic regularization. 
Removing \textsc{GCLoss} yields a smaller but noticeable drop in Macro-$F_1$ ($0.32\%$), while increasing recall and decreasing precision, suggesting that without gradient shaping the model becomes more prone to over-predict toxic instances and thus sacrifices precision. 

For toxic span extraction, \textsc{GCLoss} has a substantially larger impact than ARCL.
Removing ARCL causes a moderate degradation, with extraction $F_1$ dropping by $2.74\%$ and both recall and precision declining accordingly. In contrast, removing \textsc{GCLoss} leads to a much sharper performance drop: extraction $F_1$ decreases by $12.75\%$, accompanied by substantial reductions in both recall and precision. When CuSA is removed and the raw high-saliency tokens are used directly as gradient-boosting targets, span extraction recall drops markedly, while precision increases to a certain extent. This suggests that a model trained only with binary classification labels can still capture some relatively salient toxic expressions, but learning a broader range of toxic expressions requires additional external supervision. Finally, compared to the RoBERTa baseline, the full model improves classification Macro-$F_1$ by $0.92\%$ and boosts extraction $F_1$ by $12.82\%$, demonstrating the effectiveness of our joint training strategy.
 

\subsection{Effect of Warm-up Steps}

Since our training pipeline requires warming up the foundation model for several epochs, this section analyzes the impact of the number of warm-up steps on the final classification and extraction results.

The number of warm-up steps determines the extent to which the model learns toxicity classification solely from binary labels. As shown in Figure~\ref{fig:accuracy_curve}, too few or  too many warm-up steps lead to a decrease in the final classification accuracy.

\subsection{Gradient Attribution Faithfulness}

In NLP, the faithfulness of explanations concerns whether the highlighted evidence truly reflects the causal decision process, rather than merely correlating with its prediction. Recent work argues that faithful explanations should be grounded in counterfactual reasoning and formalizes this intuition via \emph{order-faithfulness}, showing that non-causal feature-importance methods can mis-rank evidence under confounding effects~\cite{ICLR2024_not_faithful}. Motivated by this view, we evaluate our gradient-based toxic span explanations from a causal perspective: if the extracted spans constitute genuine decision evidence, masking them should substantially reduce the model's predicted toxicity confidence.

We extract toxic spans on the COLD dataset using two RoBERTa models, both localized by BiCSE: 
(1) a baseline RoBERTa trained with binary cross-entropy loss, and 
(2) a RoBERTa trained with our full objective in Eq.~\eqref{total loss}.

Figures~\ref{confidence_drop_1} and~\ref{confidence_drop_ours} report the confidence drop, defined as the decrease in the predicted toxicity probability after masking the extracted spans. 
Compared with the BCE-only baseline, the ToxiTrace-trained model exhibits a consistently larger confidence reduction when its predicted spans are masked. 
Meanwhile, Figure~\ref{confidence_drop_0} shows that randomly masking the same number of tokens extracted by BiCSE increases the average confidence from 0.8873 to 0.9019, corresponding to an average "drop" of $-1.64\%$.

These results indicate that the extracted spans are more critical to the model's decision and therefore more faithful to its prediction.

\section{Discussion of Adaptability to Other Languages}

Our primary focus is on languages where characters are the basic units, because for word-based languages (like English), standard binary classification training often already leads to gradients concentrating on words of higher importance (even if the distribution is somewhat scattered)~\cite{kamp-etal-2023-topk}. In contrast, for character-based languages, meaning typically emerges only across multiple consecutive characters, making contiguous span extraction more critical.

Given data availability, we choose Chinese as the main target language due to the existence of relatively large-scale datasets. We will treat extending ToxiTrace to other character-based languages like Japanese and Korean as future work; if we can obtain sufficiently large datasets in these languages, we will conduct further analysis and adaptation studies.

\section{Conclusion}

We proposed ToxiTrace, which can automatically produce span-level annotations when fine-grained Chinese toxicity labels are unavailable. It further introduces a gradient-based loss to increase the model's gradient responses on fine-grained toxic spans, and aligns each sentence with its corresponding reasoning in the semantic space, achieving simultaneous improvements in classification accuracy and interpretability. In addition, we design an algorithm for extracting high-saliency tokens, which addresses the limitation of prior methods that cannot recover contiguous high-saliency spans.

\section*{Acknowledgments}

This work was supported by the Shanghai "Science and Technology Innovation Action Plan" Project (No.23511100700).

\section*{Limitations}

In real-world scenarios, some toxic speeches are "cloaked" or obfuscated, such as through the use of homophones or Pinyin replacements (as noted in CNTP). Most models experience a performance degradation when processing such obfuscated toxic information. While datasets for this exist in the Chinese domain, our current method does not specifically address these perturbations. We intend to incorporate robustness against such variations into the scope of our future research.

Our method has been developed and evaluated only on Chinese toxic content, which has unique linguistic properties (e.g., character-level tokenization and ambiguous semantic units). As a result, its effectiveness on languages with different structures remains to be verified.

\section*{Ethical Considerations}

This work addresses toxic content detection and explanation, which necessarily involves exposure to offensive and harmful language. Although the proposed method is designed to improve the faithfulness and transparency of model explanations, there is a potential risk that fine-grained toxic span extraction could be misused to reverse-engineer moderation systems or to craft adversarial toxic expressions that evade detection. To mitigate this risk, we emphasize that the proposed framework is intended for research and system auditing purposes, rather than as a fully automated content moderation solution, and should be deployed with appropriate human oversight.

All datasets used in this study (COLD, ToxiCN, and CNTP) are publicly available benchmark datasets released for academic research. They do not contain personally identifying information, and no additional data collection is conducted in this work. While the datasets do include offensive and toxic language by design, they are used solely for model training and evaluation in a controlled research setting. We do not attempt to identify, target, or profile any individuals, and no user-level or sensitive attributes are inferred or stored.

\bibliography{custom}

\clearpage

\appendix

\section{Algorithm}

The algorithm pseudo-code to extract continuous token spans.

\label{algorithm}

\begin{algorithm}[H]
    \caption{Bidirectional Salient Span Extraction}
    \label{alg:AOA}
    \renewcommand{\algorithmicrequire}{\textbf{Input:}}
    \renewcommand{\algorithmicensure}{\textbf{Output:}}
    \begin{algorithmic}[1]
        \REQUIRE Gradient sequence $G = \{g_1, g_2, \ldots, g_n\}$
        \ENSURE Salient span set $S$
        
        \STATE $\mu \leftarrow \text{Mean}(G)$
        \STATE $\tau \leftarrow \text{Median}(|g_i - g_{i-1}|)$ for $i \in [2,n]$
        \STATE
        \STATE \textbf{Function} FindCliffEnd($G, \text{start}, \mu, \tau$):
        \STATE \quad $p \leftarrow \text{start}$
        \STATE \quad \textbf{for} $i = \text{start}$ to $n$ \textbf{do}
        \STATE \quad \quad \textbf{if} $g_i > \mu$ \textbf{then} $p \leftarrow i$
        \STATE \quad \quad \textbf{if} $i \leq n - 2$ and $g_i - g_{i+1} > \tau$ \textbf{then}
        \STATE \quad \quad \quad \textbf{if} $g_{i+1} - g_{i+2} \leq \tau$ \textbf{then return} $i$
        \STATE \quad \quad \textbf{if} $g_i \leq \mu$ \textbf{then return} $p$
        \STATE \quad \textbf{return} $p$
        \STATE
        \STATE \textbf{Function} ForwardScan($G, \mu, \tau$):
        \STATE \quad $S \leftarrow \emptyset$, $i \leftarrow 2$
        \STATE \quad \textbf{while } $i \leq n$ \textbf{ do}        
        \STATE \quad \quad \textbf{if} $g_i > \mu$ and $g_i - g_{i-1} > \tau$ \textbf{then}
        \STATE \quad \quad \quad $s \leftarrow i$, $e \leftarrow \text{FindCliffEnd}(G, s, \mu, \tau)$
        \STATE \quad \quad \quad $S \leftarrow S \cup \{(s, e)\}$, $i \leftarrow e + 1$
        \STATE \quad \quad \textbf{else}
        \STATE \quad \quad \quad $i \leftarrow i + 1$
        \STATE \quad \textbf{return} $S$

        \STATE
        \STATE $S_{\text{fwd}} \leftarrow \text{ForwardScan}(G, \mu, \tau)$
        \STATE $G \leftarrow \text{Reverse(G)}$
        \STATE $S_{\text{bwd}} \leftarrow \text{Reverse}(\text{ForwardScan}(G, \mu, \tau))$
        \RETURN $\text{Merge}(S_{\text{fwd}} \cup S_{\text{bwd}})$
        
    \end{algorithmic}
\end{algorithm}

\noindent
\textbf{Threshold Computation}\quad Given a gradient sequence $G = {g_1, g_2, \ldots, g_n}$, we compute two thresholds:

Mean threshold $\mu = \text{Mean}(G)$, serving as the baseline for identifying high-gradient tokens.
Difference threshold $\tau = \text{Median}(|g_i - g_{i-1}|)$, capturing the typical magnitude of gradient transitions.

\noindent
\textbf{Start Condition}\quad A span begins at position $i$ if the gradient exhibits a steep ascent: 

$$g_i > \mu \quad \text{and} \quad g_i - g_{i-1} > \tau$$

\noindent
\textbf{Termination Condition 1 (Cliff Edge Detection)}\quad The span terminates at position $i$ when a cliff edge is detected—i.e., the current position shows a steep descent but the subsequent descent diminishes: 

$$g_i - g_{i+1} > \tau \quad \text{and} \quad g_{i+1} - g_{i+2} \leq \tau$$

This condition identifies the boundary where the gradient "cliff" ends, ensuring complete span extraction by continuing to search until the final cliff edge is found.

\noindent
\textbf{Termination Condition 2 (Fallback)}\quad If no cliff edge is detected, the span ends at the last position where $g_i > \mu$, preventing incomplete extraction when gradients decay gradually.

\noindent
\textbf{Bidirectional Scanning}\quad To capture spans that may be missed by unidirectional scanning, we perform forward scanning on $G$ and backward scanning on the reversed sequence $G'$. The final span set is obtained by merging overlapping intervals from both directions: 

$$S = \text{Merge}(S_{\text{fwd}} \cup S_{\text{bwd}})$$

This bidirectional approach ensures robust detection of salient spans regardless of their orientation within the sequence.

\section{Training Details}
\label{train-detail}

This section lists the hyperparameter settings used during training.

Warm-up stage: the RoBERTa encoder learning rate is 5e-5, and the final binary classification head learning rate is 1e-3. We use the standard AdamW optimizer with a cosine-decay learning-rate schedule. Since the classification head is randomly initialized, we adopt a larger learning rate to accelerate convergence. When extracting high-saliency cue tokens, we use the top 0.15 quantile.

Joint training stage: the RoBERTa encoder learning rate is 1e-5, and the final binary classification head learning rate is 2e-5. We use the standard AdamW optimizer with a cosine-decay learning-rate schedule. Considering that the model is primarily optimized for span extraction and binary classification, ARCL mainly serves as a semantic regularizer; therefore, we set $\lambda_{grad}=0.8$ and $\lambda_{sem}=0.5$. In the gradient loss, the PGR margin is set to 1. In the PPT loss, we set $\alpha=1.1$ and $\tau_{cap}=10$ (this parameter did not take effect in practice: it was originally introduced to prevent gradient explosion, but during training we did not observe any token gradient norm exceeding 10). For ARCL, the InfoNCE temperature is set to $\tau=0.05$ (this parameter is relatively sensitive: we tried 0.01, 0.02, 0.05, 0.1, and 0.2; temperatures that are too high or too low reduce classification performance, while their impact on span extraction is relatively small). The core training code is available at \href{https://github.com/ArdentLiby/ToxiTrace}{https://github.com/ArdentLiby/ToxiTrace}.

\section{Changes in Saliency Map}
\label{change_of_sal}

\begin{figure*}[ht]
    \centering
    \includegraphics[width=1.0\linewidth]{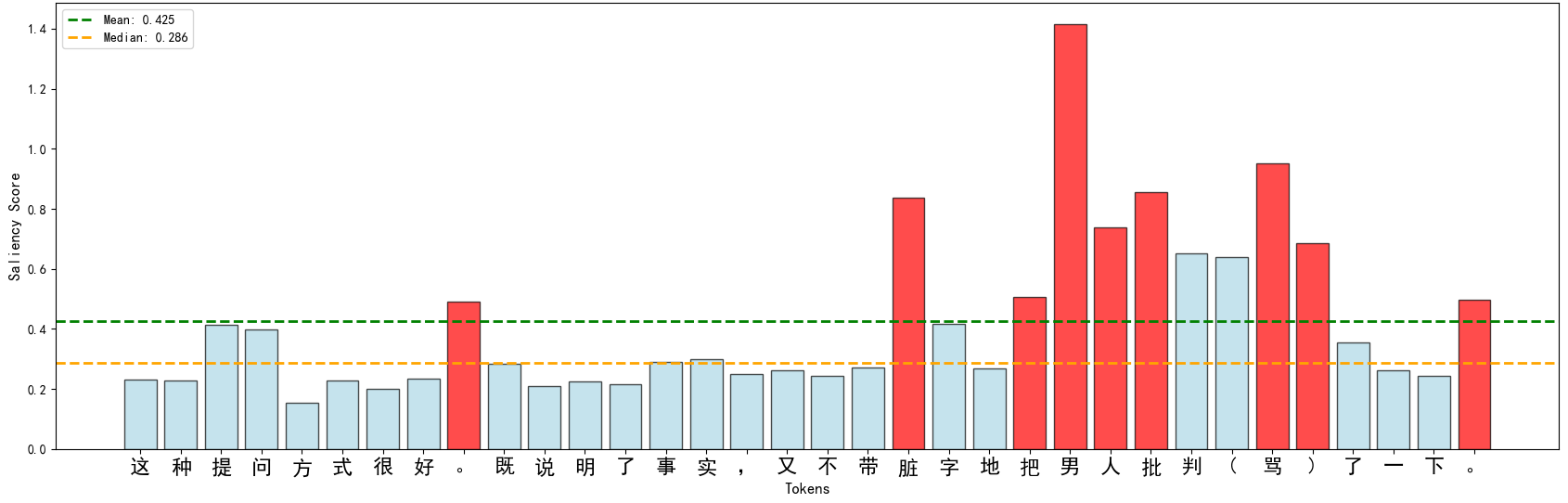} \\
    \caption{Saliency map of a RoBERTa model trained only with binary classification labels.}
    \label{harmless figure}
\end{figure*}

\begin{figure*}[ht]
    \centering
    \includegraphics[width=1.0\linewidth]{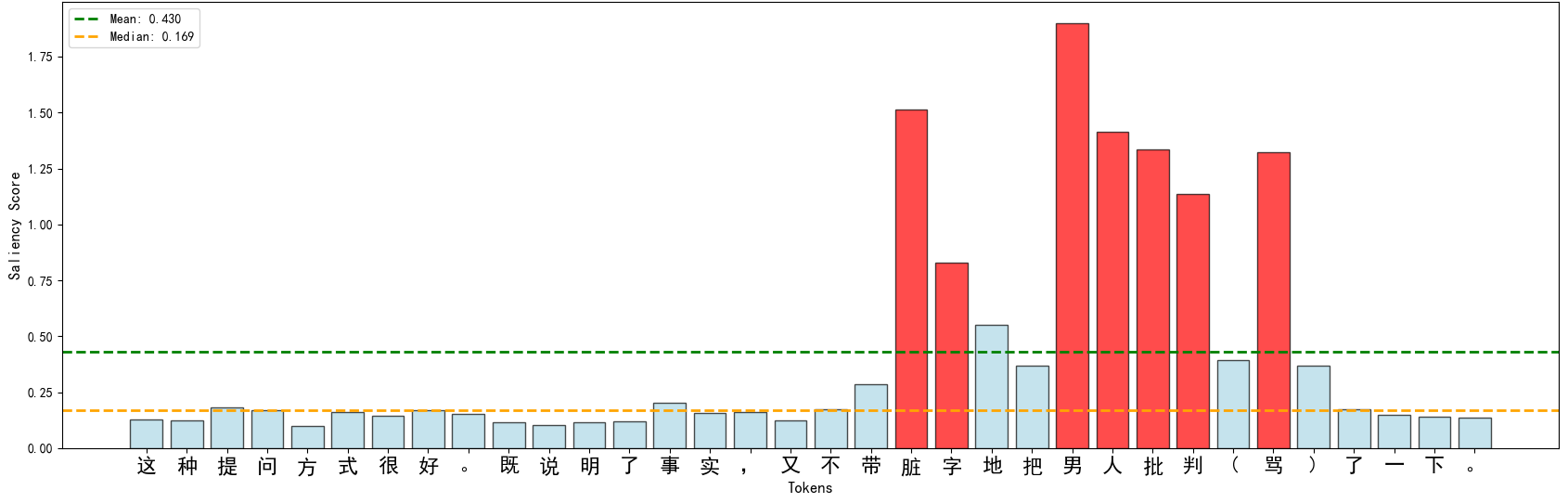} \\
    \caption{Saliency map of a RoBERTa model after training with our proposed method.}
    \label{toxic figure}
\end{figure*}

Models trained with our proposed method exhibit a pronounced shift in saliency maps, characterized by substantially higher saliency scores assigned to toxic spans. Figure~\ref{harmless figure} and~\ref{toxic figure} presents an example illustrating the saliency shift before and after training with proposed ToxiTrace, demonstrating that the model becomes more focused on truly toxic spans.


\clearpage
\onecolumn
\section{Prompt Templates for LLMs}
\label{appendix_prompt}

In this study, LLMs are employed for multiple tasks. Table~\ref{tab:Spans_Refin} presents the prompt templates used for toxic span extraction. Tables~\ref{tab:Reason_Gen_Tox} and~\ref{tab:Reason_Gen_NTox} provide the prompts designed to elicit sample-specific adversarial explanations for toxic and non-toxic instances, respectively. Table~\ref{tab:LLM_Clas} summarizes the prompts used to prompt LLMs to directly assess the toxicity of input texts.


\begin{table}[H]
\centering
\begin{tcolorbox}[
    colback=white,
    colframe=gray!120,
    width=\textwidth, 
    title=Toxic Spans Refining Prompt,
    fonttitle=\bfseries,
    rounded corners,
    boxrule=1pt,
    left=8pt,
    right=8pt,
    top=8pt,
    bottom=8pt,
    fontupper=\scriptsize
]
\setstretch{1.2}
\textbf{System:}

你是一个有害内容识别领域的专家，在你目前对"有害内容"的定义的基础之上，请对以下这个用户提供的内容进行有害内容实体的片段定位任务，实体片段尽量短且尽量包含多的有害信息（请参考后面的输出范例，所有输入句都\textbf{一定存在这样的片段}，每句话都会给出句中可疑的一些字段作为线索供参考）。

片段之间不要有重合或包含，也要注意中文隐式有害俚语。输出不需要给出任何解释、不要进行任何额外说明。输出示例（使用单行形式，不要套入 markdown 格式）：

有害实体1、有害实体2、有害实体3
\newline

\textbf{真实数据示例1：}

用户输入：确实现在的拳师天生就是坏种它们贪婪傲慢懒惰暴食淫欲嫉妒暴怒七宗罪它们全部都犯了它们欲壑难填利令智昏是畜生不如的东西

输出：拳师、是坏种、淫欲、七宗罪、都犯了、欲壑难填、畜生不如

\textbf{真实数据示例2：}

用户输入：厦门河南人和江西人开出租的很多，都挺狠的

输出：河南人、江西人、挺狠

\textbf{真实数据示例3：}

用户输入：这帮黑人不仅有枪还有毒品，一贫如洗？

输出：黑人、有枪、有毒品
\newline

\textbf{User:}

\{sentence\}

【参考线索（可能相关的重点词汇）】：\{hint\_text\}

\end{tcolorbox}
\caption{Toxic Spans Refining Prompt}
\label{tab:Spans_Refin}

\begin{tcolorbox}[
    colback=white,
    colframe=gray!120,
    width=\textwidth, 
    title=Translation of Toxic Spans Refining Prompt,
    fonttitle=\bfseries,
    rounded corners,
    boxrule=1pt,
    left=8pt,
    right=8pt,
    top=8pt,
    bottom=8pt,
    fontupper=\scriptsize
]
\setstretch{1.2}
\textbf{System:}

You are an expert in the field of toxic content recognition. Based on your current definition of "toxic content," please perform a toxic entity span extraction task on the content provided by the user below. The entity spans should be as short as possible while containing as much toxic information as possible.
\newline

(Please refer to the output examples below. All input sentences \textbf{definitely contain such fragments}, and some suspicious fields in the sentence will be given as clues for reference).
\newline

There should be no overlap or containment between fragments. Pay attention to implicit toxic slang. 
\newline
Output Requirement: Do not provide any explanation or extra description. 
\newline
Output Example (use a single line format, do not wrap in markdown):
Toxic Entity 1, Toxic Entity 2, Toxic Entity 3
\newline

\textbf{Real Data Example 1:}

User Input: Indeed, current feminists are born bad seeds; they are greedy, arrogant, lazy, gluttonous, lustful, jealous, and wrathful—committing all seven deadly sins. Their desires are insatiable, they are blinded by greed, and they are less than beasts.

Output: feminists, are bad seeds, lustful, seven deadly sins, committing all, desires are insatiable, less than beasts

\textbf{Real Data Example 2:}

User Input: There are many taxi drivers from Henan and Jiangxi in Xiamen, and they are all pretty ruthless.

Output: Henan people, Jiangxi people, pretty ruthless

\textbf{Real Data Example 3:}

User Input: These black people not only have guns but also drugs, and yet they claim to be destitute?

Output: black people, have guns, have drugs
\newline 

\textbf{User}:

\{sentence\}

【Reference Clues (Potential Key Vocabulary)】: \{hint\_text\}

\end{tcolorbox}
\caption{Translation of Toxic Spans Refining Prompt}
\label{tab:Spans_Refin_t}

\end{table}

\begin{table}[t!]
\centering
\begin{tcolorbox}[
    colback=white,
    colframe=gray!120,
    width=\textwidth, 
    title=Toxic Reasoning Generation Prompt,
    fonttitle=\bfseries,
    rounded corners,
    boxrule=1pt,
    left=8pt,
    right=8pt,
    top=8pt,
    bottom=8pt,
    fontupper=\scriptsize
]
\setstretch{1.3}

你是一个中文毒性内容理解的专家，用户将给定文本，请完全站在“这段内容有害”的主观立场上认定该内容包含对他人的攻击/威胁/贬损/歧视等“有害表达”。
\newline

请只输出一句简洁理由，不要有任何多余的内容，需要针对这句话的内容进行针对性的解释，不要输出宽泛、模糊的解释。
\newline

参考输出（仅作为示例，\textbf{并非}输出语句格式模版，但需至少包含其中提到针对该句的要素，即群体、观点、用词、可能的负面理解等）：
\newline

这段话包含对[某一群体]的[xx]性质的贬损性描述，使用了“[xxx]”这样的词汇，容易被理解为嘲讽或不尊重，也可能会被理解为人身攻击。
\newline

输入：\{sentence\}

\end{tcolorbox}
\caption{Toxic Reasoning Generation Prompt}
\label{tab:Reason_Gen_Tox}
\end{table}

\begin{table}[t!]
\centering
\begin{tcolorbox}[
    colback=white,
    colframe=gray!120,
    width=\textwidth, 
    title=Translation of Toxic Reasoning Generation Prompt,
    fonttitle=\bfseries,
    rounded corners,
    boxrule=1pt,
    left=8pt,
    right=8pt,
    top=8pt,
    bottom=8pt,
    fontupper=\scriptsize
]
\setstretch{1.3}

You are an expert in understanding toxic content. The user will provide text. Please stand completely on the subjective standpoint that "this content is toxic" and determine that the content contains "toxic expressions" such as attacks, threats, derogation, or discrimination against others.
\newline

Please output only one concise reason. Do not include any superfluous content. You need to provide a specific explanation targeting the content of this sentence; do not output broad or vague explanations.
\newline

Reference Output (This is \textbf{not} a template, but the output must include elements targeting the sentence, such as the group, viewpoint, wording, and possible negative interpretations):
\newline

This statement contains a derogatory description of the [xx] nature regarding [specific group], using vocabulary like "[xxx]", which is easily understood as mockery or disrespect, and may also be interpreted as a personal attack.
\newline

Input: \{sentence\}
\end{tcolorbox}
\caption{Translation of Toxic Reasoning Generation Prompt}
\label{tab:Reason_Gen_Tox_t}
\end{table}

\begin{table}[t!]
\centering
\begin{tcolorbox}[
    colback=white,
    colframe=gray!120,
    width=\textwidth, 
    title=Non-toxic Reasoning Generation Prompt,
    fonttitle=\bfseries,
    rounded corners,
    boxrule=1pt,
    left=8pt,
    right=8pt,
    top=8pt,
    bottom=8pt,
    fontupper=\scriptsize
]
\setstretch{1.3}

用户将给定文本，请完全站在“这段内容完全正常”的主观立场上认定该内容完全不包含对他人的攻击/威胁/贬损/歧视等任何“有害表达”。
\newline

请只输出一句简洁理由，不要有任何多余的内容，需要针对这句话的内容进行针对性的解释，不要输出宽泛、模糊的解释。
\newline

参考输出（\textbf{并非}输出语句格式模版，但需包含其中提到针对该句的要素，即群体、观点、用词、不可能产生负面理解的原因等）：
\newline

这段话针对[xx]群体发表了[客观观点]，表达一种情绪宣泄，即使使用了“[xxx]”这样的词汇，也只是语气比较重，没有贬损的意味。
\newline

输入：\{sentence\}

\end{tcolorbox}
\caption{Non-toxic Reasoning Generation Prompt}
\label{tab:Reason_Gen_NTox}
\end{table}

\begin{table}[t!]
\centering
\begin{tcolorbox}[
    colback=white,
    colframe=gray!120,
    width=\textwidth, 
    title=Translation of Toxic Spans Refining Prompt,
    fonttitle=\bfseries,
    rounded corners,
    boxrule=1pt,
    left=8pt,
    right=8pt,
    top=8pt,
    bottom=8pt,
    fontupper=\scriptsize
]
\setstretch{1.3}
The user will provide text. Please stand completely on the subjective standpoint that "this content is completely normal" and determine that the content does not contain any "toxic expressions" such as attacks, threats, derogation, or discrimination against others.
\newline

Please output only one concise reason. Do not include any superfluous content. You need to provide a specific explanation targeting the content of this sentence; do not output broad or vague explanations.
\newline

Reference Output (This is \textbf{not} a template, but the output must include elements targeting the sentence, such as the group, viewpoint, wording, and reasons why it cannot generate negative interpretations):
\newline

This statement expresses an [objective viewpoint] regarding the [xx] group, serving as an emotional outlet. Even though it uses vocabulary like "[xxx]", it is merely a strong tone and carries no derogatory implication.
\newline

Input: \{sentence\}

\end{tcolorbox}
\caption{Translation of Toxic Reasoning Generation Prompt}
\label{tab:Reason_Gen_Tox_t}
\end{table}

\begin{table}[t!]
\centering
\begin{tcolorbox}[
    colback=white,
    colframe=gray!120,
    width=\textwidth, 
    title= LLM Classification Task Instruction,
    fonttitle=\bfseries,
    rounded corners,
    boxrule=1pt,
    left=8pt,
    right=8pt,
    top=8pt,
    bottom=8pt,
    fontupper=\scriptsize
]
\setstretch{1.3}

你是一个有害内容识别领域的专家，现在有一些潜在的有害内容文本需要你来分析，可能的有害内容包括但不限于：
\newline

地域歧视：针对某些地区或该地区的人发表的整体贬低性言论；

性别对立：表达针对某一性别的贬低或歧视性言论；

种族歧视：针对特定人种发表的歧视性表达。

其余未列在上述分类中的如果有明显的攻击性及仇恨言论，也认为是有害内容。
\newline

你只需要对用户提供的内容进行判断，然后对你认为有害的内容输出一个“1”，无害内容输出“0”即可，无需进行任何其他分析，只需要输出一个“0”或“1”即可。
\newline

输入：\{sentence\}

\end{tcolorbox}
\caption{LLM Classification Task Instruction}
\label{tab:LLM_Clas}
\end{table}

\begin{table}[t!]
\centering
\begin{tcolorbox}[
    colback=white,
    colframe=gray!120,
    width=\textwidth, 
    title=Translation of LLM Classification Task Instruction,
    fonttitle=\bfseries,
    rounded corners,
    boxrule=1pt,
    left=8pt,
    right=8pt,
    top=8pt,
    bottom=8pt,
    fontupper=\scriptsize
]
\setstretch{1.3}
You are an expert in the field of toxic content recognition. There are some potentially toxic texts that need your analysis. Possible toxic content includes but is not limited to:
\newline

Regional Discrimination: Overall derogatory remarks against specific regions or people from those regions;
Gender Opposition: Expressing derogatory or discriminatory remarks against a specific gender;
Racial Discrimination: Discriminatory expressions against specific races.
\newline

If there are obvious offensive or hate speech remarks not listed in the above categories, they are also considered toxic content.
\newline

You only need to judge the content provided by the user. Output a "1" for content you consider toxic, and "0" for non-toxic content. Any other analysis is prohibited; strictly output only a "0" or "1".
\newline

Input: \{sentence\}

\end{tcolorbox}
\caption{Translation of LLM Classification Task Instruction}
\label{tab:LLM_Clas_t}
\end{table}

\end{document}